\pdfoutput=1

\documentclass[11pt]{article}

\usepackage[]{EMNLP2022}
\usepackage{graphicx}
\usepackage{times}
\usepackage{latexsym}
\usepackage{booktabs}
\usepackage{amssymb}
\usepackage{amsmath}
\usepackage{arydshln}
\usepackage{comment} 

\usepackage[encapsulated]{CJK}
\usepackage[T1]{fontenc}

\usepackage[utf8]{inputenc}

\usepackage{microtype}

\usepackage{inconsolata}

\usepackage{makecell}
\usepackage{multirow}
\title{Automatic Scene-based Topic Channel Construction System for E-Commerce}

\author{
  Peng Lin$^\ast$, Yanyan Zou{\thanks{\ \ The first two authors made equal contributions. Correspond to Yanyan Zou.}}  , Lingfei Wu, Mian Ma,  Zhuoye Ding, Bo Long \\
  JD.com, Beijing, China\\
  {\{linpeng47,zouyanyan6,lingfei.wu,mamian,dingzhuoye,bo.long\}@jd.com}  
 }

\begin{document}
	\maketitle
	\begin{abstract}
	Scene marketing that well demonstrates user interests within a certain scenario has proved effective for offline shopping.
		To conduct scene marketing for e-commerce platforms, this work presents a novel product form, scene-based topic channel which typically consists of a list of diverse products belonging to the same usage scenario and a topic title that describes the scenario with marketing words.
		As manual construction of channels is time-consuming due to billions of products as well as dynamic and diverse customers' interests, it is necessary to leverage AI techniques to automatically construct channels for certain usage scenarios and even discover novel topics.
		To be specific, we first frame the channel construction task as a two-step problem, i.e., scene-based topic generation and product clustering, and propose an E-commerce Scene-based Topic Channel construction system (i.e., ESTC) to achieve automated production, consisting of scene-based topic generation model for the e-commerce domain, product clustering on the basis of topic similarity, as well as quality control based on automatic model filtering and human screening.
		Extensive offline experiments and online A/B test validates the effectiveness of such a novel product form as well as the proposed system.
		In addition, we also introduce the experience of deploying the proposed system on a real-world e-commerce recommendation platform.
	\end{abstract}
	
	\section{Introduction}

\begin{figure*}[t]
		\centering
		\includegraphics[width=0.82\linewidth]{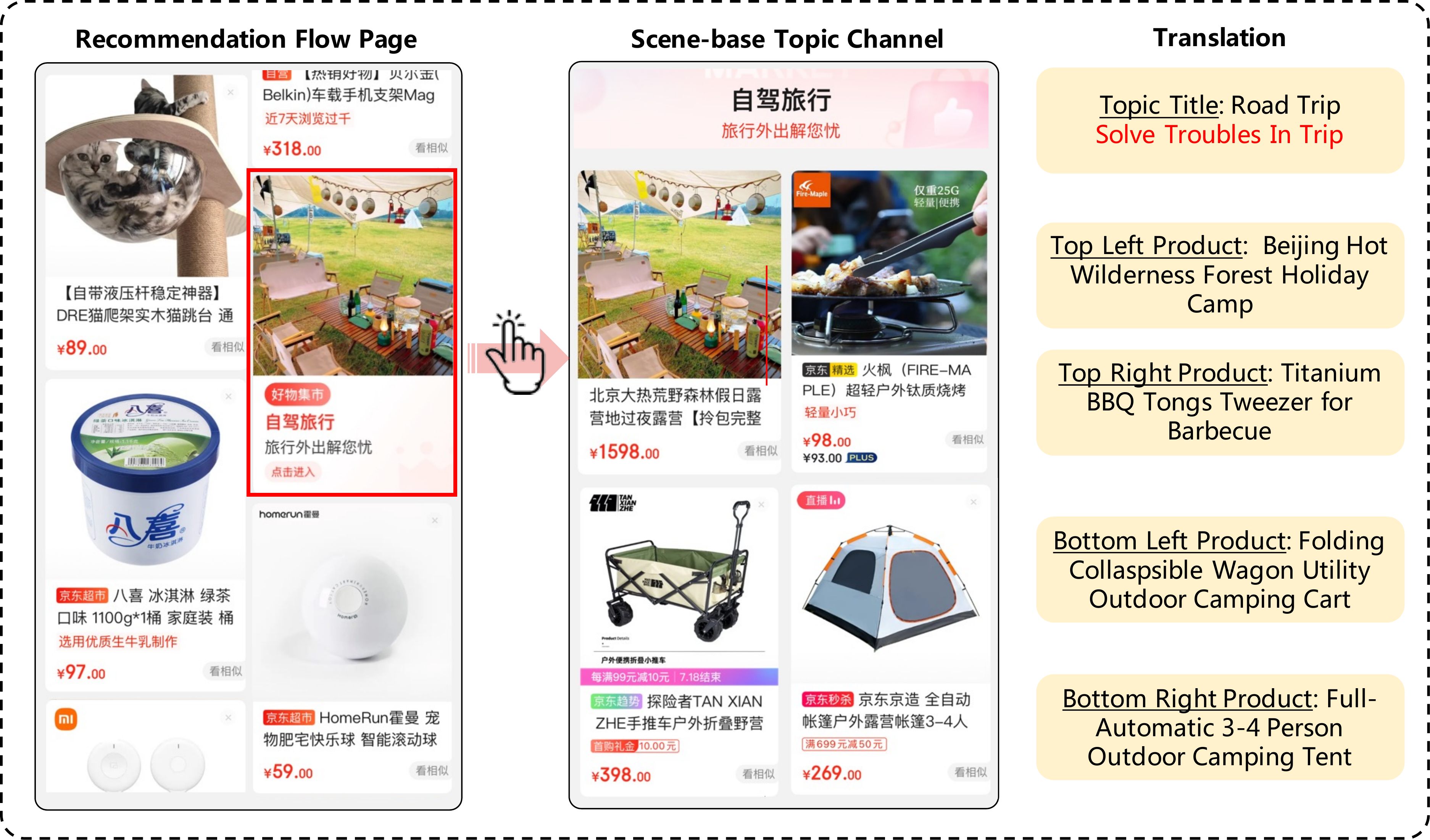}
		\caption{A screenshot of a scene-based topic channel on an e-commerce platform, with only four products due to limited space. Text with underline in the right-side ``Translation'' column are used to connect the translated words with associated parts in the topic channel. }
		\label{fig:theme_demo}
	\end{figure*}

	Recently, e-commerce platforms have become an indispensable part of people's daily life.
	Different from brick-and-mortar stores where salespersons can hold face-to-face conversations to promote products and even recommend more products related to customers' interests, most recommendation systems of e-commerce platforms, such as Taobao\footnote{\url{{https://www.taobao.com/}}}, 
	mainly display individual products in which users might be interested~\cite{zhou2018deep,zhou2019deep}, as listed in Figure~\ref{fig:theme_demo} (indicated as Recommendation Flow Page).
	Recently, scene marketing has become a new marketing mode for product promotion where particular application scenarios (i.e., scene) are created to demonstrate product functions and highlight features correspondingly~\cite{zhao2020data}, which is also paramount for e-commerce platforms to improve user experience during online shopping~\cite{Kang_2019_CVPR,fu2019constructing}. 
	A practical usage scenario of products can help users better understand product functions and features, and also allow the platform to exhibit more products that hit customer's specific interests, so that the user experience and click rate might be improved.
	However, scenes do not always help. For example, displaying all related products belonging to the same scene in the recommendation flow page might harm the user experience, since they tend to be homogeneous.

	
	To achieve scene marketing in e-commerce platforms, this work presents a novel product form, scene-based topic channel, which consists of a list of diverse products belonging to the same scenario, together with two short phrases (or sentences) as the topic title summarizing the scene.
	Exemplified by Figure \ref{fig:theme_demo}, one primary product of a channel and the associated scene topic title (highlighted with red box) are displayed in the recommendation flow page. If a user is interested in the primary product and clicks on it, the user is then redirected to the topic channel page where diverse products belonging to the same usage scenario are displayed.
	Existing ways to constructing scene-based topic channel mainly rely on expert knowledge and past experience of business operators in grouping products into different functional categories with certain scene topics~\cite{mansell2002constructing,cooke2006regional,fernandez2010ontological}.
	However, such methods are highly expensive with low efficiency and even impractical since there are billions of products in the e-commerce platforms.
	Therefore, in this work, we propose an E-commerce Scene-based Topic Channel construction system (i.e., ESTC) to automatically construct such scene-based topic channels, where the task is framed as a two-step problem, i.e., scene-based topic generation and product clustering.
	One intuitive solution to obtaining scene topics is to make use of topic models~\cite{DBLP:journals/jmlr/BleiNJ03, roberts2013structural,grootendorst2022bertopic}
	or techniques from extractive summarization~\cite{basave2014automatic,wan2016automatic},
	which are, however, restricted to assigning topics within a predefined limited candidate set, while there are often emerging scenes in the e-commerce fields.
	Thus, like \citet{DBLP:conf/sigir/AlokailiAS20}, we propose to generate scene-based topic titles for products, which allows to create novel topics not featured in the training set.

	Nevertheless, in practice, the limitation of labeled data for training (around 5000 instances) hinders the generation quality of the model.
	On the other hand, we observe that generated topic titles, describing the same scenario, might be slightly different in formulation.
	Simply grouping products based on exact string match of generated topic titles results in channels with rare products.
	To address above issues, we first develop a pre-trained model in the e-commerce field to improve generation quality.
	Then, a semantic similarity based clustering method is designed to conduct product clustering to form the channel.
	Finally, to ensure the user experience online, we further design a quality control module to strictly filter out undesired channels, such as inconsistent topic titles, or  channels with irrelevant topic-product pairs.
	Our contributions are summarized as follows:
	\begin{itemize}
		\setlength\itemsep{0.0001em} 
		
		\item A topic generation model in e-commerce field is proposed to generate scene-based topic titles for products, which is flexible to produce topics for emerging products and allows the system to discover novel scene topics.
		\item A semantic similarity based clustering method is designed to aggregate products with similar topic titles and form scene-based channels, which is able to improve the product diversity. 
		
		\item A quality control module is designed to ensure the quality of the artificially constructed channels before they are released online.
		
		\item  We introduce the overall architecture of the deployed system where the ESTC has been successfully implemented into a real-world e-commerce platform.
		
		
		\item To the best of our knowledge, this is the first work on automatically constructing scene-based topic channel for scene marketing in e-commerce platforms.
	\end{itemize}

	\section{Proposed Method}
	The development of the proposed ESTC system consists of three main parts, including scene-based topic generation for each product, scene-based product clustering to aggregate products with similar topic titles, as well as the quality control module to ensure the quality of AI-generated channels.
	We also include a simple data augmentation module to discover weakly supervised data in order to improve the diversity of generated topic titles.
	
	

	\subsection{Scene-based Topic Generation}
	
	
		In this work, we propose to generate the scene-based topic titles for each product.
	To be specific, given input information $X= (x_1, x_2, \dots, x_{|X|})$ of a product $P$, including product's title $T$,  a set of attributes $A$ and side information $O$ obtained through optical character recognition techniques, paired with scene-based topic title $Y=(y_1, y_2, \dots, y_{|Y|})$, we aim to learn model parameters $\theta$ and estimate the conditional probability:
	\begin{equation*}
		\vspace{-0mm}
		P(Y|X;\theta) = \prod_{t=1}^{|Y|} p(y_t|y_{<t};X;\theta)
	\end{equation*}
	where $y_{<t}$ stands for all tokens in a scene title before position $t$ (i.e., $y_{<t}=(y_1, y_2, \dots, y_{t-1})$).
	
		\paragraph{Pretraining with E-commerce Corpus} Pre-trained  models~\cite{radford2019language,DBLP:conf/naacl/DevlinCLT19,lewis-etal-2020-bart,2020t5,zou2020pre,xue-etal-2021-mt5} have proved effective in many downstream tasks, however, most of which are developed on English corpora from general domains, such as news articles, books, stories and web text.
	In our scenario, we aim to produce topic titles in Chinese that summarize certain usage scenarios of products.
	Therefore, a model is required to understand the products through its associated information (such as title, semi-structured attributes) and generate scene-based topic titles, where we argue that the model should learn knowledge from e-commerce fields and thus propose to further pre-train models in domain~\cite{DBLP:conf/acl/GururanganMSLBD20}.
	Specifically, besides the product title, attribute set as well as side information, we also collect the corresponding advertising copywriting of products from e-commerce platforms for the second phase of pre-training.
	We adopt the UniLM~\cite{dong2019unified} with BERT initialization as backbone structure.
	
	Recall that the product attributes $A$ is a set without fixed order.
	We observe that input containing same attributes yet in different orders might results in different outputs.
	On the other hand, UniLM is an encoder-decoder shared architecture.
	To reinforce both the understanding and generation ability of no-order input information, in addition to the original pre-training objectives of UniLM, we also propose two objectives to adapt the target domain:
	\begin{itemize}
		\setlength\itemsep{0.0001em} 
		\item Consistency Classification:  Given a product title-attributes pair, this task aims to classify if the two refer to the same product. For the positive example, the attributes and the title describe the same product and attributes are randomly concatenated as a sequence to introduce disorder noises. For the negative example, we randomly select attributes from another different product.
		\item Sentence Reordering:  We split the product copywriting into pieces according to marks (such as comma and period).
		Such pieces are then shuffled and concatenated as a new text sequence.
		The model takes the shuffled sequence as input and learns to generate the original copywriting.
	\end{itemize}
	After the second phase of pre-training in the target e-commerce domain, we fine-tune the pre-trained model on the scene-based topic generation dataset.

    	\subsection{Scene-based Product Clustering}
		One intuitive solution to constructing a scene-based topic channel is to group products with exactly the same generated topic titles.
		However, we observe there exists channels with similar topic titles, each of which merely contains several products, while we expect one channel has diverse products to ensure user experience.
	Therefore, we design a clustering module to aggregate products with semantically similar topic titles.
	

	\paragraph{Topic Encoding}
	To better learn scene-based topic representations and distinguish different topic titles, we take all topic titles from training set as input and employ the SimCSE~\cite{DBLP:conf/emnlp/GaoYC21} to further fine-tune the e-commerce pre-trained UniLM model in an unsupervised fashion.
	The embeddings of the last layer are used as the initialization for product clustering.
	

	\paragraph{Product Clustering} 
	This module aims to group products with semantically similar topic titles into a cluster, in other words, a product list for a channel.
	Since we do not have prior knowledge of how many topic clusters the topic generation model would produce, we adopt the hierarchical clustering~\cite{sahoo2006incremental} where the number of kernels is not required.
	To be specific, we adopt the bottom-up version, namely Agglomerative Nesting, which treats each sample as a leaf node and uses an iterative method for aggregation. 
	In each iteration, two nodes with the highest similarity score are merged to form a new parent node. 
	The iterative process stops when the shortest distance among all nodes is greater than a preset threshold.
	It is worthy noting each cluster might align with multiple topic titles and a list of products.
	The display order of products within a channel is decided by recommendation strategies, which is not focus of this work.
	

			\begin{figure}[tb]
		\centering
		\includegraphics[width=\linewidth]{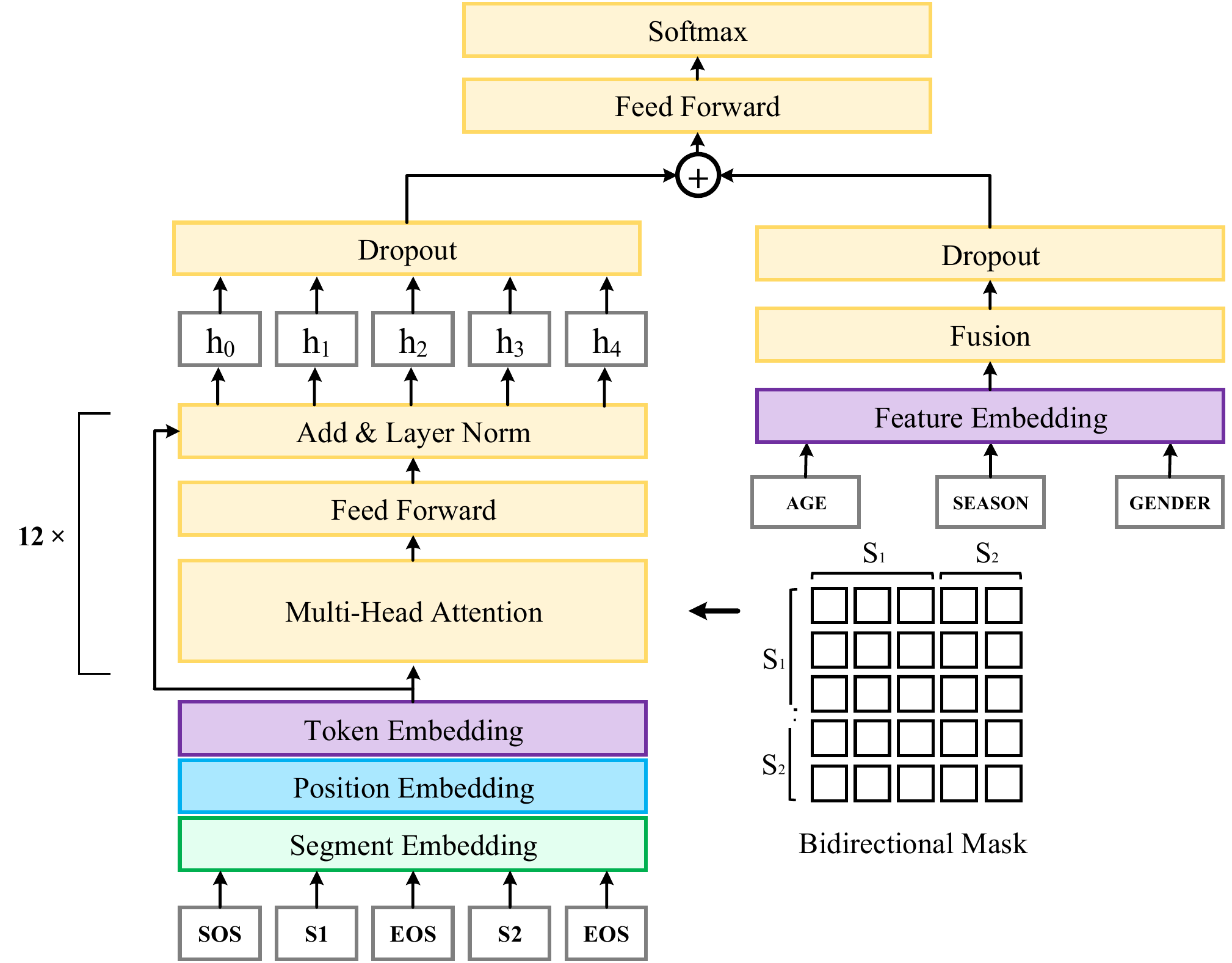}
		\caption{Correlation scoring model structure.}
		\label{fig:cls_filter_model}
	\end{figure}
	
	\subsection{Quality Control}
	Although our method can generate good-quality channels most of the time, there is still possibility that the generated channels might not be accurate: 1) the generated topic title is semantically incoherent; 2) the topic title and associated products are not related according to the product usage scenarios.
	Thus, in order to alleviate above issues and ensure a reasonably good experience online, we design two modules, sentence coherence and correlation scoring models, to remove unexpected samples.
	%
	%
	%

	\paragraph{Topic Coherence Model}
	We empirically observe that the generated topic titles might suffer coherent issues, like repetition and incompletion.
	Thus, we design a topic coherence model to classify if a generated topic title is coherent.
	To be specific, we resort the e-commerce UniLM model with a softmax layer to classification.
	During training, we treat the online published topic titles as positive examples.
	The negative ones are synthesized:
	\begin{itemize}
		\setlength\itemsep{0.00001em} 
		\item  {Samples with repetition:} For a positive example of topic title, each unigram and bigram is selected and repeated for one or two times with equal probability.
		\item  {Incomplete samples}:  We randomly remove the last two bigram or unigram tokens of a positive topic title.
	\end{itemize}
We randomly select above synthesized samples to make the number of negative examples equal to the size of positive examples.
Recall that a cluster might have multiple topic title candidates, the one with highest coherence score by the topic coherence model is used as final topic title.
If all title candidates are classified as incoherent, then we simply remove such a cluster.
After this module, each cluster is a scene-based topic channel with a list of products belonging to the same scene as well as a topic title summarizing the scene.
	
	\begin{figure}[t]
		\centering
		\includegraphics[width=\linewidth]{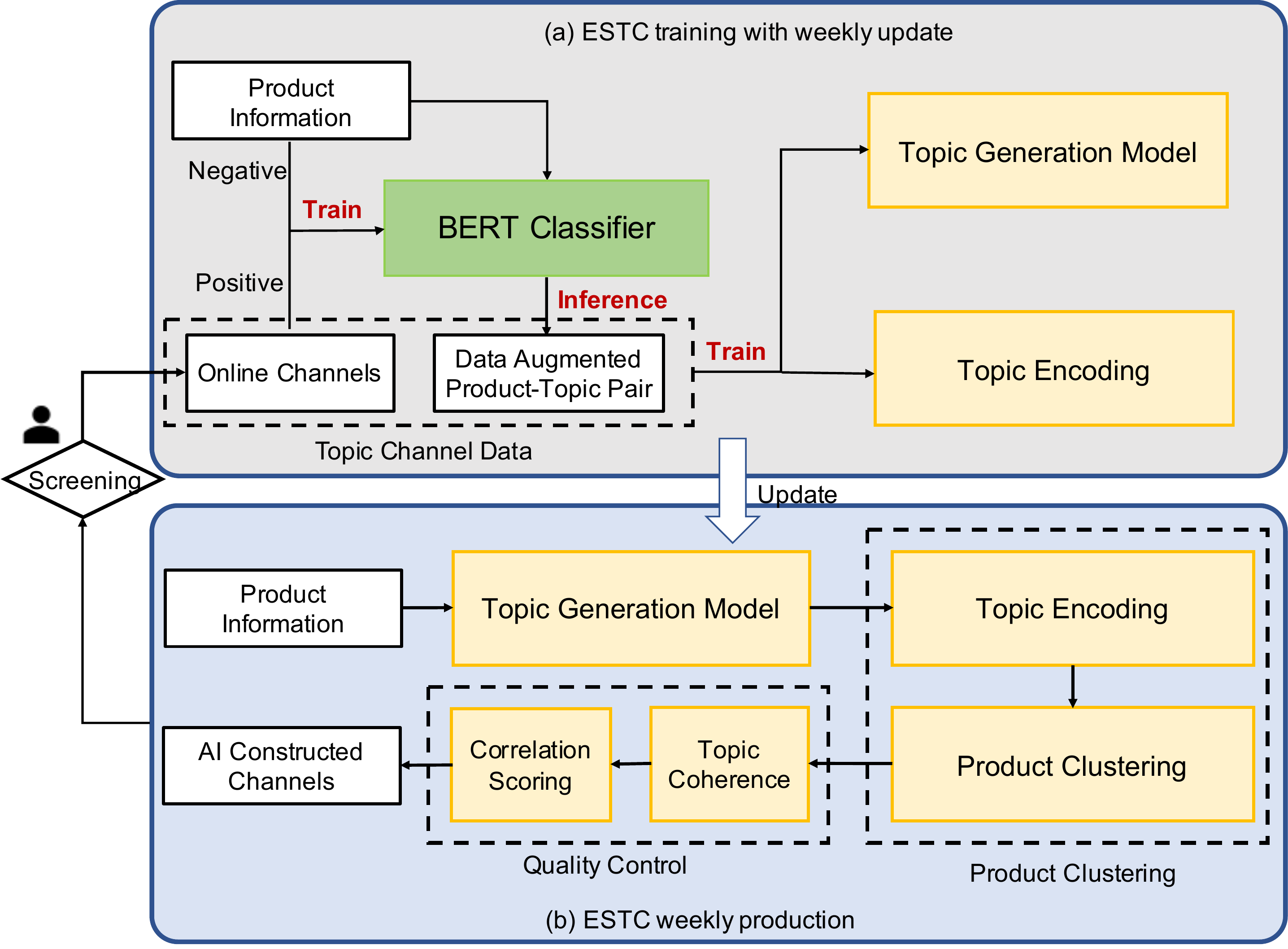}
		\caption{ESTC deployment on e-commerce platform.}
		\label{fig:demploy}
	\end{figure}

	\paragraph{Correlation Scoring Model}
	We design another binary classification model, i.e., correlation scoring model, to identify if the topic title and products are scene-based related.
	As illustrated in Figure~\ref{fig:cls_filter_model}, the e-commerce UniLM model takes as input the product information of a single product $X$ as well as the generated topic title $Y$ and determine whether they are related by the relevant scene.
	For better learning the product usage scenario, we also take into account the product profile information, such as age, season, and gender profiles, and employ a feed-forward layer to encode such features.
	
	Likewise, product-topic title pairs from online published topic channels are considered as positive examples.
	The negative samples are obtained by randomly selecting mismatched product-topic title pairs.
	As a result, the number of negative examples is the same as the positive ones.
	
	For each constructed channel, we use this model to check each product-topic title pair and remove products that are unrelated to the topic.	
	\subsection{Data Augmentation}
	Initially, the online existing (i.e., human-created) topic channels are quite limited which might hinder the model performance.
	Moreover, we would like to construct novel channels.
	Thus, we propose a  UniLM-based binary classification model to discover more and diverse product-topic title candidate pairs.
	To be specific, the existing online product-topic title pairs are considered as positive examples.
	Similar to \citet{zhang2022automatic}, a product with its side information $O$ from product detail images are considered as negative examples.
	After the classification model is trained, we use such a model to further extract more data for training. 
	Negative examples with high probability scores are augmented into the training set.
	
    \section{Deployment}
	
	We have successfully deployed the proposed ESTC system on a real-world e-commerce platform.
	Figure~\ref{fig:demploy} demonstrates the workflow of the deployed system with weekly update.
	Firstly, the data augmentation module is utilized to augment existing online channels.
	The augmented data is then used to train the topic generation and encoding models.
	Since there are thousands of millions products online, we weekly update the model and re-construct the channels to discover novel scene-based topic channels.
	To ensure a proper user experience, human screening is necessary before publishing channels online.

	\section{Experiment}
\begin{table}[t]
		\centering
			\begin{tabular}{ccccc}
				\toprule
				\textbf{Dataset} & \textbf{\#PT} & \textbf{\#T} & \textbf{IL} & \textbf{OL}\\
				\midrule
				Human & 177,412 & 5,186 & 69.34 & 13.44 \\
				Mined & 111,572 & 82,834 & 74.21 & 12.54 \\
				\bottomrule
			\end{tabular}
			\caption{The statistics of topic generation dataset. \#PT denotes the number of product-topic pairs, \#T denotes the number of topic titles, IL denotes the average length of input product information sequence and OL denotes the average length of topic titles.}
			\label{tab:dataset_status}
	\end{table}
	
		\begin{table*}[t]
		\centering

			\scalebox{0.85}{\begin{tabular}{lccccccc}
				\toprule
				\textbf{Model} & \textbf{SacreBLEU} & \textbf{ROUGE-1} & \textbf{ROUGE-2} & \textbf{ROUGE-L} & \textbf{BLEU} & \textbf{METEOR} & \textbf{DR(\%)}\\
				\midrule
				BART & 1.92 & 7.50 & 1.02 & 7.01 & 3.20 & 8.63 & 1.09 \\
				UniLM-BERT  & 2.05 & 7.87 & 1.11 & 7.42 & 3.45 & 8.70 & 0.87 \\ 
				E-commerce UniLM   & 2.08 & \textbf{8.01} & 1.12 & \textbf{7.56}  & 3.47 & \textbf{8.78} & 0.88  \\ 
				E-commerce UniLM + DA & \textbf{2.17} & 7.68 & \textbf{1.21} & 7.36 & \textbf{3.68} & 8.70 & \textbf{12.07} \\
				\bottomrule
			\end{tabular}}
			\caption{The results of different topic generation model.}
			\label{tab:topic_generation_exp}
	\end{table*}

	\subsection{Topic Generation Results}
	\label{sec:tg_test}
	 \paragraph{Data Collection}
	The data for developing scene-based topic generation model consists of two sources: existing online channels (including human-created) and augmented samples, collected from a publicly available online e-commerce platform, JD.com\footnote{\url{https://www.jd.com/}}. 
	For the scene-based topic channels, we collected online channels from the product form ``Goods List'' from the platform, which were reviewed by human.
	
	We also leveraged the optical character recognition (OCR) and classification techniques to extract key information about the product from product detail images. Firstly, texts are extracted from the images. Then, the extracted texts are ranked in descending order of their importance and relevance using the classification model. Finally, highly ranked texts are selected and merged as the final OCR input of products for topic generation. 
	
	In the end, we have 5,186 topic titles created by human and 82,834 topic title candidates from product side information.
	We further split the whole dataset into training, validation and test set with a ratio of 80\%:10\%:10\%. 
	The online channels are considered as ground-truth.
	Details are listed in Table~\ref{tab:dataset_status}.
	Moreover, we constructed product-OCR text (i.e., side information) pairs for the data augmentation module.

	\paragraph{Comparison} We use SacreBLEU \cite{post2018call}, ROUGE \cite{lin2004rouge}, BLEU \cite{papineni2002bleu}, and METEOR \cite{lavie2004significance} to measure quality of outputs by different generation models.
	We also design a new metric, difference rate (i.e., DR), to measure the novelty of generated topics, which is the ratio of the number of novel topics (i.e., not appearing in the training set) and the total number of generated ones.
	We consider publicly available models pre-trained  on Chinese corpus as baselines, including BART~\cite{shao2021cpt} 
	and UniLM with BERT initialization. 
	As listed in Table~\ref{tab:topic_generation_exp}, our E-commerce UniLM model achieves best performance for most evaluation metrics.
	With augmented data (denoted as +DA), the performance of our model is further improved with more novel topic titles produced, which shows the effectiveness of the data augmentation module.

	\subsection{Product Clustering Results}
	\paragraph{Dataset} The clustering module works in an unsupervised fashion, while labeled data is still required for model evaluation.
	We manually create a data set for clustering evaluation, containing 65 different topic title samples, belonging to 18 groups.

	\paragraph{Metrics} We adopt the distance-based Silhouette Coefficient~\cite{rousseeuw1987silhouettes}	to evaluate the performance of topic clustering.
	To investigate how well a clustering matches reference partitions of the test data, we further design two metrics.
	
	For each topic sample $i$ from cluster $j$, the precision score is calculated as:
	\begin{equation}
		\centering
		P_{i,j} = \frac{TP_{i,j}}{N_j}
	\end{equation}
	where $TP_{i,j}$ denotes the number of correctly grouped topic $i$ in cluster $j$, $N_j$ is the number of samples in cluster $j$.
	Similarly, the recall score is calculated as:
		\begin{equation}
		\centering
		R_{i,j} = \frac{TP_{i,j}}{T_i}
	\end{equation}
	 where $T_i$ is the total number of topic $i$ found across all clusters.
	
	The F1-measure score is computed as the harmonic mean of precision and recall.

		\paragraph{Comparison} We compare different sentence embedding-based clustering methods, including bag of words (i.e., B.O.W), Word2Vec~\cite{mikolov2013efficient}, BERT as well as our E-commerce UniLM model.
		As listed in Table~\ref{tab:cluster_test_app}, models with SimCSE achieve better clustering performance.
		Furthermore, we compare two clustering method, K-means and hierarchical clustering methods, where the initial embedding are taken from different models.
The hierarchical clustering with SimCSE enhanced e-commerce UniLM model achieves best performance.
It is worthy noting that, the Silhouette score is not consistent with our designed metric scores.
		We practically observed that higher F1 scores indicate better clustering results for topics.

	\subsection{Quality Control Results}
	We also conducted human evaluation to investigate the effectiveness of each module for quality control, where for each setting, 1000 constructed channels are presented for human screening and report overall acceptance rate that is the ratio of the validated channels and the all candidates.
	As listed in Table \ref{tab:ablation}, considering both topic coherence and correlation scoring modules results in the highest acceptance rate, demonstrating strengths of quality control module.


%

\begin{table*}[htb]
	\centering
	\scalebox{0.95}{\begin{tabular}{lcccccccc}
			\toprule 
			\textbf{Model}&	\multicolumn{4}{c}{\textbf{Kmeans}}&	\multicolumn{4}{c}{\textbf{HC}}  \\
			& Silhouette& Recall &Precision & F1&Silhouette& Recall&	 Precision & F1\\
			\midrule 
			B.O.W &   0.221 & 81.9 & 74.3 & 72.4 &0.264 & 90.8 & 88.7 & 88.0\\
			Word2Vec &   0.191 & 82.3 & 80.3 & 78.7 &0.220 & 88.3 & 86.5 & 85.3\\
			BERT & 0.150 & 68.3 &  64.8 & 62.7 &0.200 & 75.7 & 74.9 & 73.7\\
			\quad +SimCSE &   \textbf{0.262} & \textbf{90.0} & \textbf{86.2} & \textbf{86.6} &\textbf{0.283} & 88.9 & 89.0 & 88.1\\
			E-commerce UniLM  & 0.210 & 73.5 & 69.0 & 68.4 &  0.248 & 72.3 & 68.7 & 68.0\\
			\quad +SimCSE & 0.248 & 84.5 &  83.1 & 81.9 &   \textbf{0.283} & \textbf{96.4} &   \textbf{96.0} & \textbf{95.7}\\    
			\bottomrule 
	\end{tabular}}
	\caption{The performance of different topic encoding models and different clustering models.}
	\label{tab:cluster_test_app}
\end{table*}

	\subsection{Online A/B Test}

	To demonstrate the payoff generated by ESTC system, a standard A/B testing is conducted to evaluate the benefit of deploying scene-based topic channels on an e-commerce mobile app.
	After launching such a new product form, the Click-Through Rate (CTR) is improved by 3.20\%, compared to the one without scene-based topic channels, which shows the values of AI-generated scene-based topic channels.
	We note that the comparison between human-created and AI-generated channels is difficult to fairly determine, since there are many factors mattering the online performance, such as recommendation strategies of products within channels.

	More details about generated samples are  included in Appendix.

	\section{Lessons Learned During Deployment}
Several lessons we have learned during model deployment could be beneficial for other like-minded practitioners who wish to deploy cutting-edge AI technologies into real-world applications, such as the importance of real-world data quality and business understandings.
\begin{itemize}
	\item Data quality matters model performance. Besides the model capacity, the quality of training data is of paramount importance. The cleaning procedures of raw data (e.g., removing poor samples from training set and specifying important attributes) plays a critical role in model development. 
	\item Business understandings and logic advance AI model launching. The AI constructed scene-based topic channels are not fool proof. Thus, in order to ensure a reasonably good user experience, post-processing, based on insightful business understandings and logic, of AI constructed channels in the production platform is necessary to filter out any inconsistent or low-quality contents.
\end{itemize}


	\section{Related Work}
	
	Previous studies~\cite{lau2011automatic,bhatia2016automatic,mei2007automatic} on topic mining mainly first retrieve candidate topic labels from reference corpora and then conduct topic ranking to select the best topic label.
	\citet{lau2010best} simply take a word from a top-N terms as the topic label.
	Knowledge bases are also adopted to retrieve topic labels by matching candidate topic words to knowledge concepts~\cite{magatti2009automatic,hulpus2013unsupervised}.
	Techniques from extractive summarization have also been used for topic extraction~\cite{basave2014automatic,wan2016automatic}, which typically extract summary sentences from the input text related to topics.
	Recent years have witnessed neural networks are successfully leveraged to improve performance of topic modeling techniques, such as incorporating neural embeddings into existing LDA-like models~\cite{bianchi2021cross,thompson2020topic}, as well as the clustering embedding based approaches~\cite{sia2020tired,angelov2020top2vec,grootendorst2022bertopic}.
	A potential limitation of such methods is that the topic labels are within a predefined limited candidate set, while there are often emerging scenes in the e-commerce fields.
	Therefore, similar to  \citet{DBLP:conf/sigir/AlokailiAS20},  we design a pre-trained model in e-commerce domain to generate scene-based topic titles, which allows to generate novel topics not featured in training set.

	\begin{table}[t]
	\centering
	\scalebox{0.85}{\begin{tabular}{lccc}
			\toprule 
			{Architecture}  & Acceptance Rate (\%)\\
			\midrule 
			ESTC w/o Quality Control  & 51.6\\
			\quad + Topic Coherence  & 65.6 \\
			\quad + Correlation Scoring  & 60.6 \\
			\quad + Both & 75.0\\
			\bottomrule 
	\end{tabular}}
	\caption{Human evaluation for quality control.}
	\label{tab:ablation}
	\end{table}
	Natural language processing techniques have been widely used in e-commerce fields to improve user experience, including automatic product copywriting generation~\cite{zhang2022automatic,wang2022interactive}, online product review generation~\cite{fan2019product,liu2021multi} and question generation~\cite{gao2020learning,deng2020opinion}. 
	Differently, we propose to leverage natural language generation and clustering techniques to automatically construct scene-based topic channels, which, to the best of our knowledge, is novel.

	\section{Conclusion}
	This work aims to automatically construct scene-based topic channels.
	According to the understanding of business requirements, we propose to first generate topic titles for each product following by conducting product clustering to form a channel and design a novel framework, consisting of topic generation, product clustering and post-processing modules.
	The extensive offline experiments and online A/B test have demonstrated the effectiveness of the proposed approach.
	Incorporating user behaviors (e.g., click preference) into channel construction processes is worthy investigating in future. For example, the generated title and the clustered product are personalized.

	\section{Ethical Considerations}
	The data used in this work are collected from a publicly available online e-commerce platform, where the collection process is consistent with the terms of use, intellectual property and privacy rights of the platform.
	The annotated data for clustering evaluation are constructed by authors, where the process is fair for all models.
	Please note that no private user data was used during data collection process.
	
	The proposed ESTC system can be deployed on various e-commerce platforms where the scene marketing is required.
	On the other hand, the display style (or the product form) can be changed according to the practical needs, where the proposed system can provide products that belong to same usage scenarios.
	
	Moreover, the AI constructed channels are not fool proof. Thus, as we discussed in the paper, in order to ensure the users can have a reasonably good experience, quality control and human screening of AI generated channels in the production platform is necessary to filter out any inconsistent or low-quality content. 

	\section{Acknowledgements}
	We thank all the anonymous reviewers for their constructive comments.

	\bibliography{anthology,custom}
	\bibliographystyle{acl_natbib}
	
	\appendix

	\section{Experiment Settings}
	\label{sec:appendix}
	
	\subsection{The Hyper-parameters of Scene-based Topic Generation Model}
	
    In this section, we introduce the detailed setting of proposed Scene-based Topic Generation. To generate the scene-based topic titles for each product, we design a topic generation model based on UniLM, the input sequence and output sequence are encoded by the same attention module with different attention masks. The model is a 12-layer transformer with multi-head attentions. During training, the learning rate is 0.00007, the warmup proportion is set to 0.2 and the batch size is 1024. The detailed hyper-parameters are listed in Table~\ref{tab:hp_PTGME}. The rest of the parameters are set by default.
	
	\begin{table}[htb]
	    \centering
	    \scalebox{0.9}{\begin{tabular}{lc}
	        \toprule 
	         hyper-parameters &  value \\
	         \midrule
	         learning\_rate & 0.00007 \\
	         warmup\_proportion & 0.2 \\
	         batch\_size & 1024 \\
	         max\_input\_length & 120 \\ 
	         max\_output\_length & 20\\
	         beam\_size & 4 \\
	         embedding\_size & 768 \\
	         hidden\_dropout\_prob & 0.1\\
	         hidden\_size &  768 \\ 
	         layer\_norm\_eps & 1e-12 \\
	         max\_position\_embeddings & 250 \\
	         num\_attention\_heads & 3 \\
	         num\_hidden\_layers & 12\\
	         activation & "gelu"\\
	         vocab\_size & 21128 \\
	         \bottomrule
	    \end{tabular}}
	    \vspace{-2mm}
	    \caption{The detailed hyper-parameters of architecture of E-commerce UniLM.}
	    \vspace{-3mm}
	    \label{tab:hp_PTGME}
	\end{table}
	
	\subsection{The Hyper-parameters of Topic Enocoding Model}
	
    The topic encoding model encodes topic texts into vector features of specified dimensions, which facilitates clustering by common numerical clustering models. In this section, we introduce the detailed setting of proposed theme encoding model.  We first obtain 700k topic data by the inference results of the E-commerce UniLM model. Then we employ SimCSE\footnote{\url{https://github.com/princeton-nlp/SimCSE}} to fine-tune the second pre-trained UniLM model in the e-commerce domain. The backbone model is a 12-layer transformer. The learning rate is 0.00003 and the batch size is 64. The rest of the parameters are set by default.
	
	\begin{table}[htb]
	    \centering
	    \scalebox{0.9}{\begin{tabular}{lc}
	        \toprule 
	         hyper-parameters &  value \\
	         \midrule
	          num\_train\_epochs & 4 \\
	          max\_len & 32 \\
              train\_batch\_size &64 \\
              learning\_rate &3e-5 \\
              max\_seq\_length &32 \\
              evaluation\_strategy& steps \\
              pooler\_type &cls \\
              temp &0.05 \\
	         \bottomrule
	    \end{tabular}}
	    \vspace{-2mm}
	    \caption{The detailed hyper-parameters of architecture of topic encoding.}
	    \vspace{-3mm}
	    \label{tab:hp_TC}
	\end{table}
	
	\subsection{The Hyper-parameters of Topic Coherence Model}
	
	Topic Coherence Model is 12-layer transformer with a feed-forward network and a softmax layer to distinguish whether the input topic is coherent.  The learning rate is 0.00005 and the batch size is 2048. The rest of the parameters are set by default.
	
	\begin{table}[htb]
	    \centering
	    \scalebox{0.9}{\begin{tabular}{lc}
	        \toprule 
	         hyper-parameters &  value \\
	         \midrule
	         learning\_rate & 0.00005 \\
	         warmup\_proportion & 0.1 \\
	         batch\_size & 2048 \\
	         max\_len & 32 \\
	         embedding\_size & 768 \\
	         hidden\_dropout\_prob & 0.1\\
	         hidden\_size &  768 \\ 
	         layer\_norm\_eps & 1e-12 \\
	         max\_position\_embeddings & 250 \\
	         num\_attention\_heads & 3 \\
	         num\_hidden\_layers & 12\\
	         activation & "gelu"\\
	         vocab\_size & 21128 \\
	         \bottomrule
	    \end{tabular}}
	    \vspace{-2mm}
	    \caption{The detailed hyper-parameters of architecture of topic coherence model.}
	    \vspace{-3mm}
	    \label{tab:hp_tcm}
	\end{table}
	
	\subsection{The Hyper-parameters of Correlation Scoring Model}
	To filter out bad cases where topic title is not suitable for product usage scenarios, we design another binary classification model, i.e., correlation scoring model, to identify if the topic title and products are scene-based related.  We concatenate the product description information $X$ and topic title $Y$ as the input of UniLM
	For better learning the product usage scenario, we also take into account the product profile information, such as age, season, and gender profiles, and employ a embedding layer and a feed-forward layer to encode such features. The detailed hyper-parameters of correlation scoring model are listed in Table~\ref{tab:hp_csm}. The rest of the parameters are set by default.
	
	\begin{table}[htb]
	    \centering
	    \scalebox{0.9}{\begin{tabular}{lc}
	        \toprule 
	         hyper-parameters &  value \\
	         \midrule
	         learning\_rate & 0.00005 \\
	         warmup\_proportion & 0.1 \\
	         batch\_size & 2048 \\
	         max\_len & 158 \\
	         feature\_embedding\_size & 300 \\
	         feature\_fusion\_size & 300 \\
	         feature\_vocab\_size & 13 \\
	         embedding\_size & 768 \\
	         hidden\_dropout\_prob & 0.1\\
	         hidden\_size &  768 \\ 
	         layer\_norm\_eps & 1e-12 \\
	         max\_position\_embeddings & 250 \\
	         num\_attention\_heads & 3 \\
	         num\_hidden\_layers & 12\\
	         activation & "gelu"\\
	         vocab\_size & 21128 \\
	         \bottomrule
	    \end{tabular}}
	    \vspace{-2mm}
	    \caption{The detailed hyper-parameters of architecture of correlation scoring model.}
	    \vspace{-3mm}
	    \label{tab:hp_csm}
	\end{table}

	\section{The Generated Scene-based Topic}
	
	In this subsection, we show some example topics generated by topic generation models, as well as examples of topics generated by the entire system, as shown in Table~\ref{tab:case_tgm} and Table~\ref{tab:case_tc}, respectively.
	
\begin{CJK*}{UTF8}{gbsn}
	\begin{table*}[htb]
	    \centering
	    \scalebox{0.82}{\begin{tabular}{l|l}
	        \toprule
	        Input &  Generated Topic\\
	        \midrule
	        创意烟灰缸生日送礼送男朋友 &   
	        送 男 友 好物\ @\ 为 爱 精 挑 细 选 \\
            Creative ashtray birthday gift for boyfriend & Gifts For Boyfriends @ Carefully Selected For Love\\
            \midrule
	        夏季宽松休闲翻领男上衣 & 
	        精选t恤\ @\ 夏日清凉出行 \\
	        Summer Loose Casual Lapel Men's Top  & Selection Of T-shirts\ @\ Summer Cool Outting \\
	        \midrule
	        网红款渔夫帽 & 
	        防晒合集\ @\ 清凉防晒一夏 \\
	        Web celebrity's fisherman hat & Sun Protection Collection\ @\ All Summer Cool Sun Protection \\
	        \midrule
	        龙井2022新茶绿茶茶礼盒装 &
	        女婿必买\ @\ 教你一招搞定老丈人 \\
	        Longjing 2022 new tea green tea gift box & Son-in-law Must Buy @ Teach You A Trick To Get Father-in-law \\
	        \bottomrule
	    \end{tabular}}
	    \caption{The generated scene-based topic titles by the topic generation model. We use \ @\ to separate two phrases of the topic title.}
	    \label{tab:case_tgm}
	\end{table*}
\end{CJK*}

\begin{CJK*}{UTF8}{gbsn}
	\begin{table*}[htb]
	    \centering
	    \scalebox{0.82}{\begin{tabular}{cc}
	        \toprule
	        Generated Topic &  Product List\\
	        \midrule
	        \multirow{4}{*}{\makecell{初生好礼\ @\ 虎娃新生儿礼物\\Newborn Gift\ @ Tiger Baby Newborn Gift}} & \makecell{尿裤尿不湿学步裤吸湿透气\\ Moisture absorbent breathable diaper toddler pants}\\
	        & \makecell{初生宝宝幼儿浴巾被子防惊跳睡袋 \\ Newborn baby bath towel quilt anti-startle sleeping bag} \\
	        & \makecell{婴儿记忆棉乳胶枕头枕芯 \\Baby memory foam latex pillow} \\
	        & \makecell{婴儿配方奶粉2段850克 \\ Infant formula milk powder 2 stage 850g} \\
	        \midrule
	        \multirow{4}{*}{\makecell{快乐露营\ @\ 露营运动欢乐时光 \\ Happy Camping\ @\ Camping Sports Happy Hour}} & \makecell{露营灯强光手电筒帐篷灯\\ Camping lights glare flashlight tent lights}\\
	        & \makecell{户外折叠桌椅 便携式野外可折叠野餐桌子\\ Portable outdoor folding picnic table} \\
	        & \makecell{登山露营保暖加宽双人户外棉睡袋 \\Mountaineering camping warm widening double outdoor cotton sleeping bag} \\
	        & \makecell{大空间防风3-4人三秒速开全自动速搭帐篷 \\ Large space windproof 3-4 people three seconds to open fully automatic tent}\\
	        \midrule
	        \multirow{4}{*}{\makecell{尽情挥洒汗水\ @\ 是兄弟一起上球场 \\ Sports Sweat\ @\ On The Court With Your Brother}} & \makecell{高帮板鞋男子经典运动休闲鞋篮球文化鞋 \\ High-top sneakers men's classic sneaker, basketball culture shoes} \\
	        & \makecell{男装梭织运动长裤运动服男 \\ Men's woven sports trousers sportswear} \\
	        & \makecell{简约经典训练系列男子圆领套头休闲百搭卫衣 \\ Simple and classic training series casual all-match sweatshirt} \\
	        & \makecell{针织五分裤男透气舒适夏季短裤男运动裤子 \\ Knitted 1/2 pants men's breathable and comfortable summer running workout joggers} \\
	        \bottomrule
	    \end{tabular}}
	    \caption{The generated scene-based topic channel of ESTC system. We use \ @\ to separate two phrases of the topic title.}
	    \label{tab:case_tc}
	\end{table*}
\end{CJK*}

\section{Examples of Scene Marketing}
\begin{figure*}
		\centering
		\includegraphics[width=\linewidth]{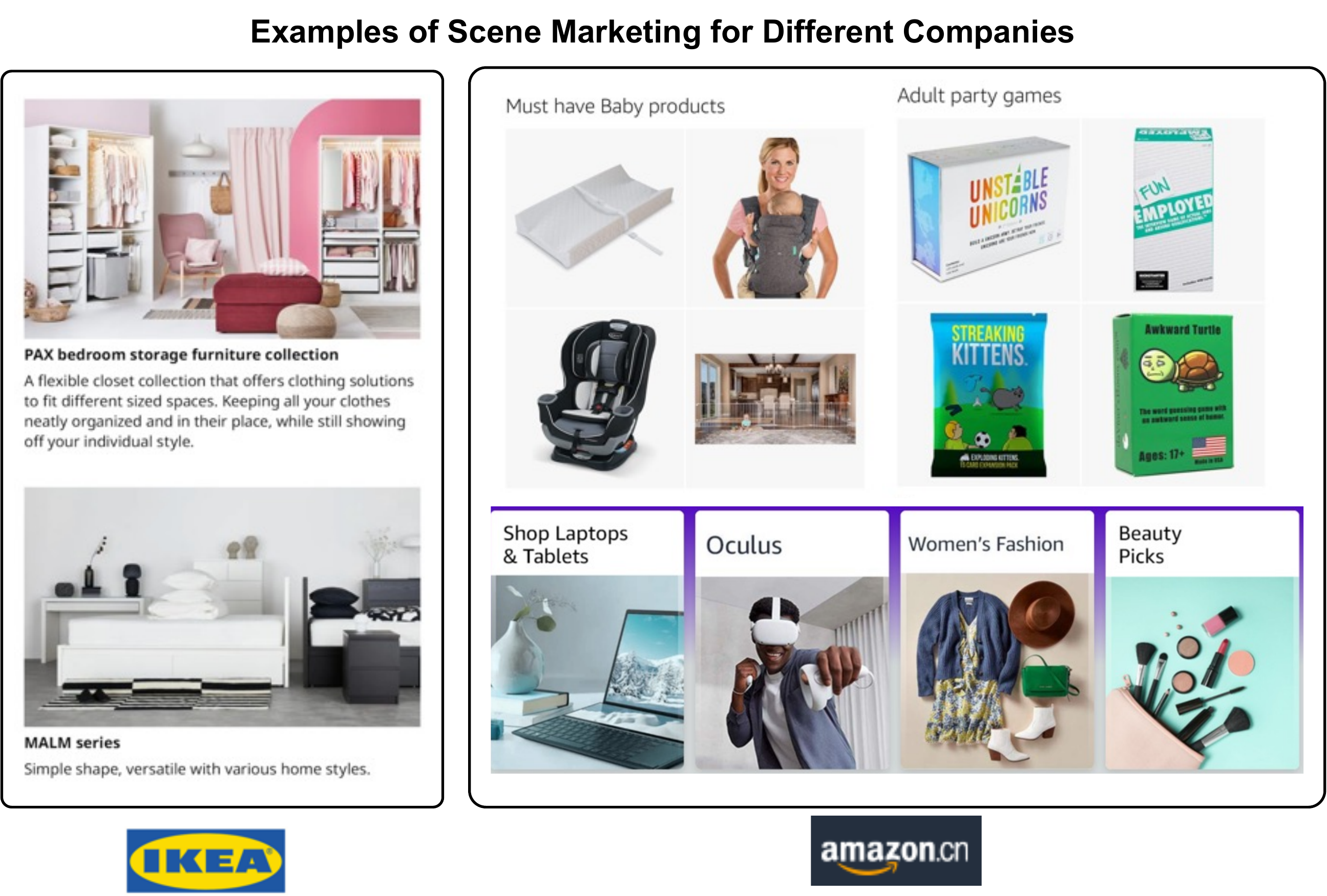}
		\caption{Examples of scene marketing for different companies.}
		\label{fig:exs_other_company}
	\end{figure*}

In recent years, many companies have begun to use scene marketing to promote products, as shown in Figure~\ref{fig:exs_other_company}, which are scene marketing for IKEA\footnote{\url{https://www.ikea.com/}} and Amazon\footnote{\url{https://www.amazon.com/}}. IKEA combines different types of furniture in a scene room to highlight key attributes of furniture, such as storage and simple shape. Amazon also exhibits the functional scenarios for products, like babysitting and party games. Such scene marketing can help consumers understand the functions and features of products, which may improve user experience and product conversion rates.

\end{document}